\documentclass[letterpaper]{article} 
\usepackage{aaai25}  
\usepackage{times}  
\usepackage{helvet}  
\usepackage{courier}  
\usepackage[hyphens]{url}  
\usepackage{graphicx} 
\usepackage{natbib}  
\usepackage{caption} 
\usepackage{algorithm}
\usepackage{algorithmic}

\usepackage{multirow, multicol}
\usepackage{makecell}
\usepackage{amssymb}
\usepackage{amsmath}
\usepackage{xcolor}

\usepackage{newfloat}
\usepackage{listings}
\DeclareCaptionStyle{ruled}{labelfont=normalfont,labelsep=colon,strut=off} 
\floatstyle{ruled}
\newfloat{listing}{tb}{lst}{}
\floatname{listing}{Listing}
\title{Discovering Conceptual Knowledge with Analytic Ontology Templates for Articulated Objects}
\author{
    Jianhua Sun\equalcontrib,
    Yuxuan Li\equalcontrib,
    Longfei Xu\footnotemark[2],
    Jiude Wei\footnote{These authors contributed equally.},
    Liang Chai,
    Cewu Lu\thanks{Corresponding Author.}
}
\affiliations{
    Shanghai Jiao Tong University
}

\begin{document}

\nocopyright

\maketitle

\begin{abstract}
  Human cognition can leverage fundamental conceptual knowledge, like geometric and kinematic ones, to appropriately perceive, comprehend and interact with novel objects. Motivated by this finding, we aim to endow machine intelligence with an analogous capability through performing at the conceptual level, in order to understand and then interact with articulated objects, especially for those in novel categories, which is challenging due to the intricate geometric structures and diverse joint types of articulated objects. To achieve this goal, we propose Analytic Ontology Template (AOT), a parameterized and differentiable program description of generalized conceptual ontologies. A baseline approach called AOTNet driven by AOTs is designed accordingly to equip intelligent agents with these generalized concepts, and then empower the agents to effectively discover the conceptual knowledge on the structure and affordance of articulated objects. The AOT-driven approach yields benefits in three key perspectives: i) enabling concept-level understanding of articulated objects without relying on any real training data, ii) providing analytic structure information, and iii) introducing rich affordance information indicating proper ways of interaction. We conduct exhaustive experiments and the results demonstrate the superiority of our approach in understanding and then interacting with articulated objects. 
\end{abstract}

\section{Introduction}

Articulated objects~\cite{xiang2020sapien,liu2022akb}, composed of rigid segments interconnected by joints that enable translation and rotation movements, play an important role in daily life. Learning articulated objects brings essential significance in a wide range of research area, including computer vision, robotics and embodied AI. Due to the intricate geometric structures and diverse joint types in articulated objects~\cite{xiang2020sapien,mo2021where2act}, it is challenging to interact with unseen articulated objects for machine intelligence, especially for those in novel categories.

On the other hand, we humans also come into contact with a large amount and variety of objects in our daily lives. Particularly, even for objects in novel categories that an individual has never seen before, \textit{i.e.} with no prior knowledge at either object or part level, studies from cognitive and brain science~\cite{carey2001infants,scholl1999explaining,leslie1998indexing,piaget1955child,ullman2000high,biederman1987recognition,hummel1992dynamic} show that human intelligence is still able to find the essence of their geometric structures and further rationally interact with them, relying on fundamental conceptual knowledge. This finding establishes a possible way for machine intelligence to interact with novel articulated objects from the conceptual perspective. However, little previous work has delved into this area in the data-driven era. 

\begin{figure}[tb!]
\centering
\includegraphics[width=0.47\textwidth]{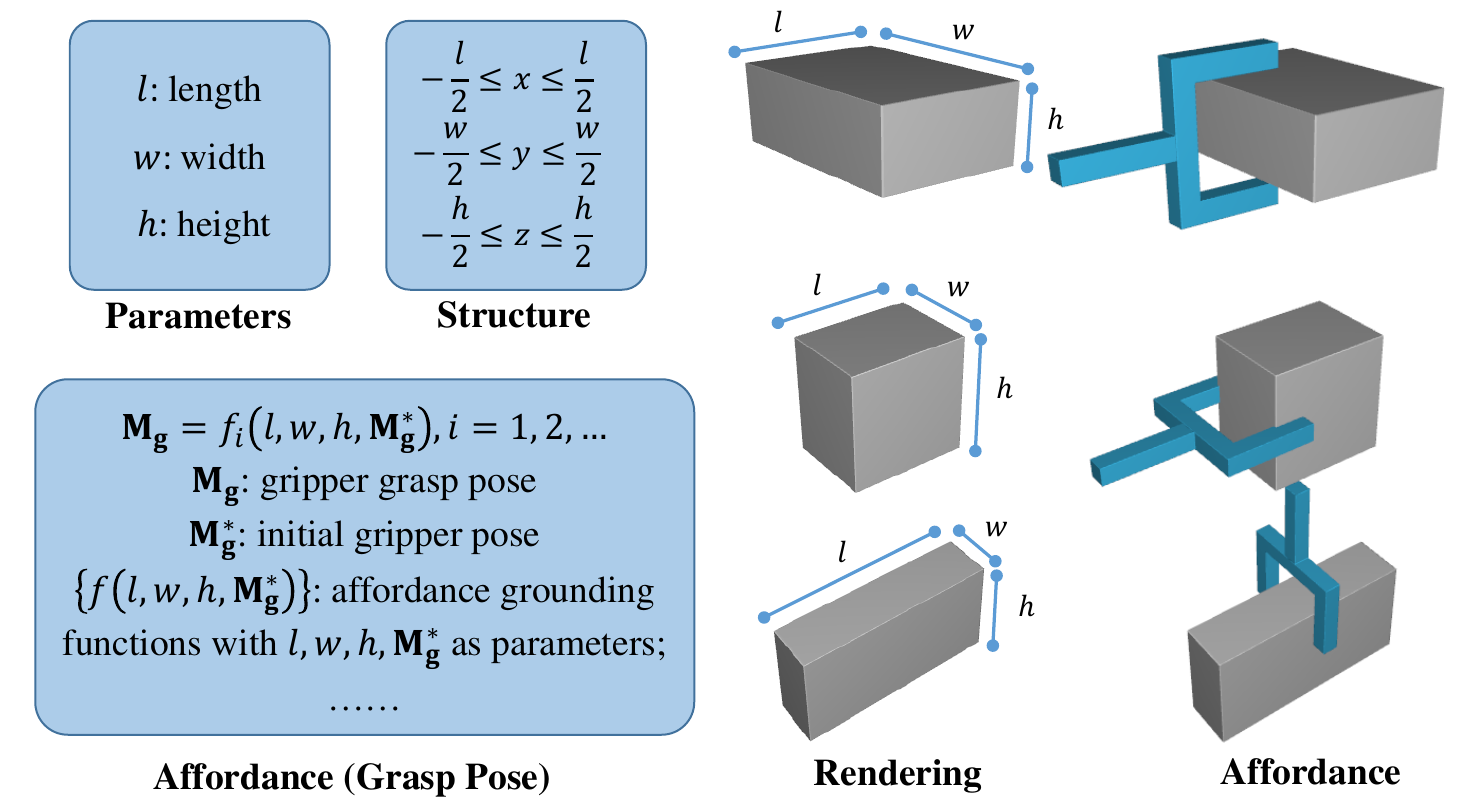}
\caption{A brief schematic of AOT with \textit{cuboid} as an example. Its structure delineates the cuboid shape, its parameters determine the size and aspect ratios, its affordances can include possible grasp poses and its renderer draws specific cuboid instances in 3D space. Here the poses $\mathbf{M}_\mathbf{g}$ and $\mathbf{M}_\mathbf{g}^*$ are in the form of affine transformation matrices.}
\label{fig:aot_1}
\end{figure}

Motivated by this idea, we seek to equip intelligent agents with fundamental and generalized concepts, allowing them to discover conceptual knowledge on articulated objects, and finally enabling them to effectively understand and properly interact with such objects. Particularly, these concepts may include geometric ones like \textit{cuboid} or \textit{ring} and kinematic ones like \textit{revolute} or \textit{prismatic}. Achieving this goal entails two key questions to answer: i) how to describe a conceptual ontology for machine intelligence and ii) how to leverage the conceptual descriptions to discover conceptual knowledge about the structure and affordance of articulated objects. 

In this paper, we present Analytic Ontology Template (AOT) to answer the first question. AOT is a template description of a conceptual ontology in a program-like style, consisting of four main components: 
\begin{itemize}
    \item \textbf{Structure}: A set of differentiable mathematical expressions to describe the intrinsic basis of the ontology.
    \item \textbf{Parameter}: The value of variables in the structure description to identify specific instances of this template.
    \item \textbf{Affordance}: Some generalized knowledge of this ontology about where and how to interact with it.
    \item \textbf{Renderer}: A tool for rendering instances of this ontology template to data of certain formats (\textit{e.g.} point clouds, meshes).
\end{itemize}

Fig.~\ref{fig:aot_1} demonstrates a brief schematic of AOT. 

\begin{figure}[tb!]
\centering
\includegraphics[width=0.47\textwidth]{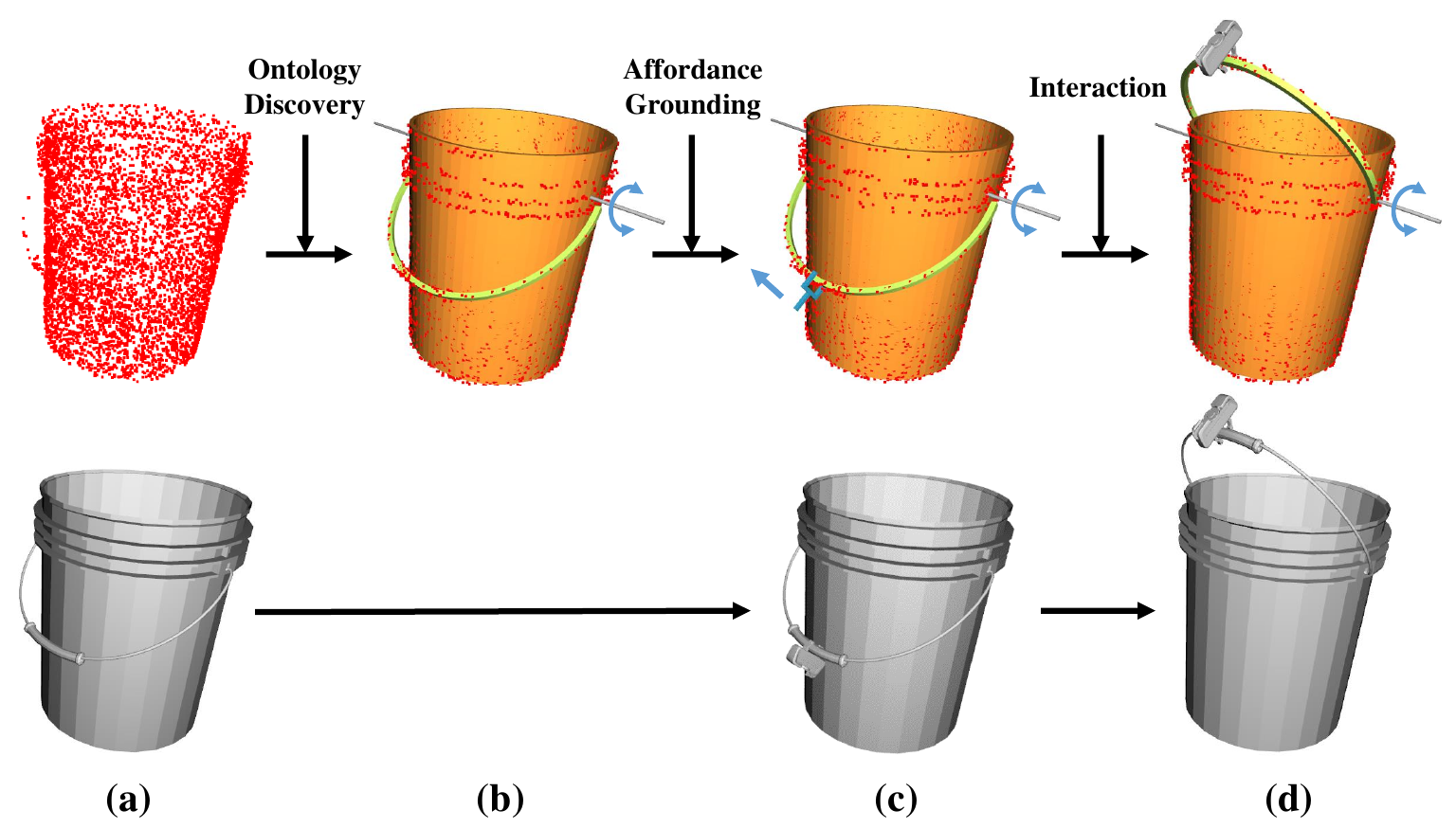}
\caption{A brief illustration of the AOTNet workflow by articulated concept discovery (a-c). The top row refers to the processing pipeline in AOTNet, and the bottom row refers to the state of the real articulated object in corresponding steps.}
\label{fig:workflow}
\end{figure}

The analytical nature of AOT stems from the mathematical expressions that comprise it, exhibiting two significant features. i) Parameterized. By varying the parameters, countless instances of an ontology template can be created. In turn, a specific instance can be resolved into parameters of an ontology template. ii) Differentiable. This enables AOTs to be directly adopted in neural networks.

Regarding the second question, we investigate it in the context of a task where an agent is required to change a novel articulated object from its initial state to a target final state. The entire process, to which we refer as articulated concept discovery for simplicity, involves the discovery of both the geometric and kinematic structure concepts of the articulated part, as well as the identification of affordance concepts pertaining to where and how to properly interact with it. The success rate of an intelligent agent on this task can directly reflect the quality of the structure and affordance concepts discovery.

Coupled with the architecture and features of AOT, we propose a baseline approach called AOTNet powered by a set of AOTs to interact with articulated objects by articulated concept discovery. Fig.~\ref{fig:workflow} gives a brief illustration of the AOTNet workflow. It first discovers the conceptual ontologies on the target, which is carried out by identifying the ontology template type and the corresponding parameters of AOT instances. This step discovers the conceptual essence of the target, \textit{e.g.} the concept \textit{ring} for the handle in Fig.~\ref{fig:workflow}. Neural networks involved in this process can be trained with synthetic data obtained by rendering AOT instances with varying parameters. The above process gives analytic estimations of the articulation structure in terms of conceptual ontology. Then, by analytically grounding the affordances of each ontology onto the target, an agent is able to appropriately interact with it guided by the knowledge. 

In light of the above points, our AOT-driven approach benefits in understanding and interacting with articulated objects from the following perspectives. i) It generalizes well across novel articulated object categories and does not rely on real training data, due to the strong generalization capability of AOTs at the conceptual level. This holds significance since abundant high quality 3D articulated data~\cite{xiang2020sapien,liu2022akb} are expensive and labor-intensive to acquire and annotate, and it is difficult to comprehensively collect data to cover the diverse categories of articulated objects in the real world. ii) It can provide analytic articulation information via conceptual ontologies with mathematically defined structures and specific parameters. iii) Affordance knowledge can be precisely grounded which brings valuable guidance for accurate and controllable interaction. To show the superiority of our AOT-driven approach, we conduct exhaustive experiments on PartNet-Mobility dataset with SAPIEN environment~\cite{xiang2020sapien}. 

In summary, the main contributions of this work are:
\begin{itemize}
    \item We propose Analytic Ontology Template (AOT) to describe conceptual ontologies for machine intelligence. It is capable of analytically describing both geometric and kinematic concepts along with affordances while remaining fully differentiable, allowing it to be incorporated into neural networks.
    \item We introduce an AOT-driven baseline, AOTNet, to equip intelligent agents with AOTs to discover conceptual knowledge about the structure and affordance on articulated objects. To the best of our knowledge, AOTNet is the first method to operate at the conceptual level, and demonstrates the effectiveness and benefits of interacting with novel articulated objects at this level.
    \item We evaluate our approach on hundreds of articulated objects across a wide range of novel categories. Through the remarkable results, we first demonstrate the possibility for machine intelligence to mimic this kind of human capability, \textit{i.e.} leveraging generalized concepts to appropriately understand and then interact with a novel object.
\end{itemize}

\section{Analytic Ontology Templates}
\label{sec:AOT}

In this section, we present Analytic Ontology Templates, \textit{a.k.a.} AOT, to describe generalized geometric and kinematic conceptual ontologies for machine intelligence. We start by introducing the basic design philosophy in Sec.~\ref{subsec:design phi}, which helps to better understand the motivation behind the specific design. Then, in Sec.~\ref{subsec:aot arch} we describe the detailed architecture of AOT, as well as some particular examples. Finally, several discussions on AOT are presented in Sec.~\ref{subsec:aot discussion}.

\begin{figure*}[tb!]
\centering
\includegraphics[width=\textwidth]{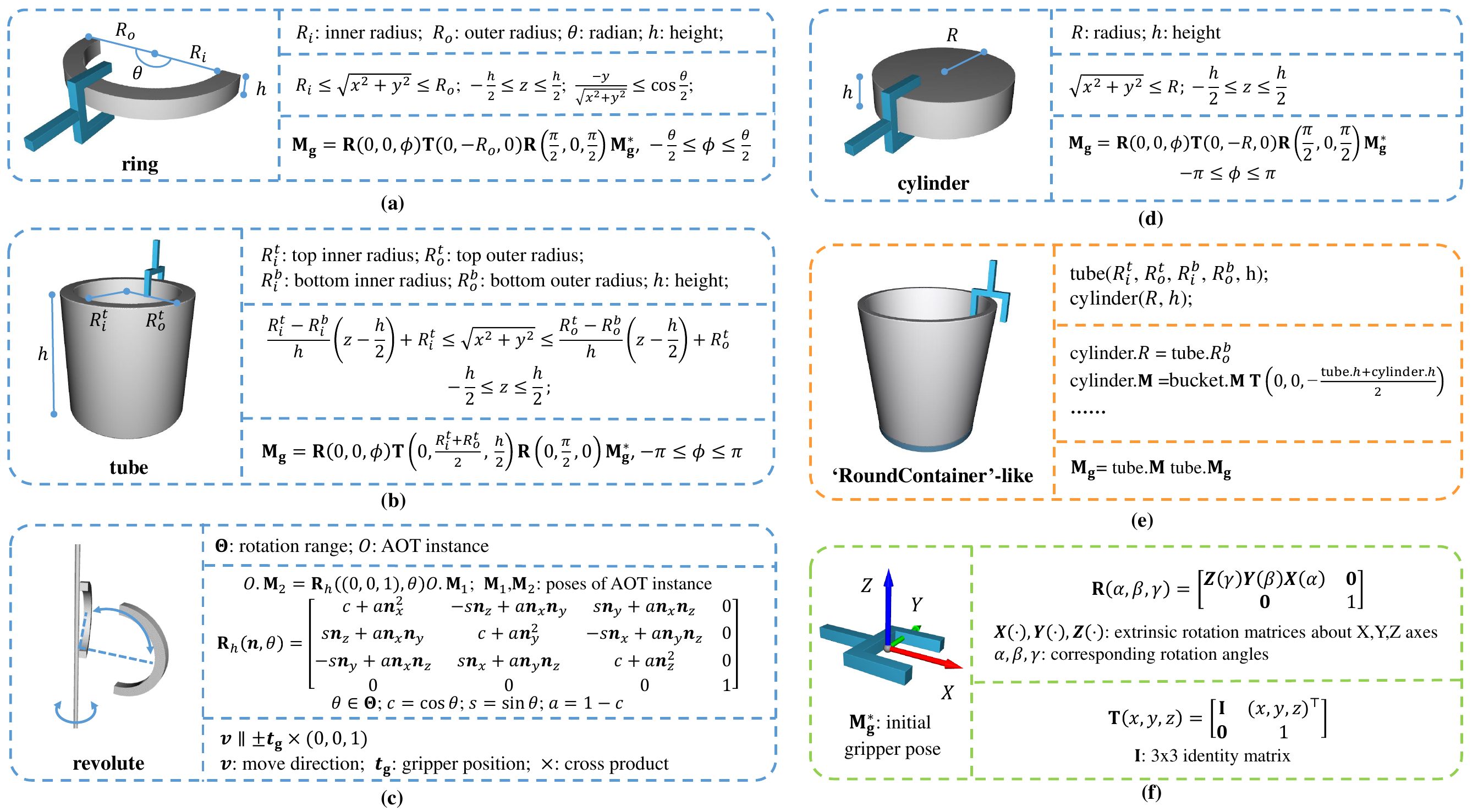}
\caption{Specific examples of basic geometric (a,b,d), kinematic (c) and composite (e) ontologies, where (a-d) are defined from scratch and (e) is built upon existing ones. Each block in (a-e) is a collection of the AOT name and rendering (left), parameters (top-right), structure (mid-right) and affordances (bottom-right, partial). (f) includes definitions of viewing angle and initial gripper pose $\mathbf{M}_\mathbf{g}^*$ (left), rotation matrix \textbf{R} and translation matrix \textbf{T} for poses (right).
Note that i) the AOT name is just a referent (like a template id) of a concept and bears no semantic meanings; ii) the definitions of basic and composite AOTs are shown in the form of mathematical expressions and pseudo codes for easy understanding, \textit{please refer to the supplementary material for examples in Python scripts} about how to define basic ones from scratch and composite ones by inheritance in an object-oriented programming fashion; iii) common parameters such as position and rotation in world coordinates are not shown for a clear view, please see Sec.~\ref{subsec:aot discussion} for details. }
\label{fig:aot_example}
\end{figure*}

\subsection{Design Philosophy} 
\label{subsec:design phi}

As the name suggests, the design philosophy of AOT covers two major points: i) the essential components of a program template and ii) the analytic nature. 

Regarding the first point, it is desirable for a template to provide adequate valid information for a conceptual ontology. Its core components should consist of a detailed delineation of the ontology basis, along with associated parameters that represent diverse variations. Then, the template ought to be incorporated with adequate knowledge of affordances to facilitate interaction, \textit{e.g.} grasp poses. Besides, a function is necessary to draw instances of the template in 3D space, \textit{i.e.} render a template instance to data, for practical use such as network training, inference and visualization.

For the second point, the analytic nature implies that the template should be comprised of mathematical analytical expressions. An analytic description of a concept provides valuable knowledge for understanding its structure and facilitating interaction. Take the concept of \textit{ring}, which frequently appears in the design of handles, as an example, by discovering the \textit{ring} concept on a handle as an analytic description, we can identify its size and pose, as well as the detailed parameters for the orientation of its hinge. Further, the affordance knowledge like grasp poses of it can be grounded according to the parameters mathematically, which helps intelligent agents conduct more precise and controllable interactions. Moreover, by ensuring all mathematical expressions being differentiable, AOTs can be incorporated into neural networks for training and optimization.

\subsection{Architecture}
\label{subsec:aot arch}

As a program template, we introduce class-like scripts to design AOTs, which facilitate easy extension as they enable a new AOT to be defined by inheritance through composition of existing ones in an object-oriented programming fashion. To satisfy the two points mentioned in Sec.~\ref{subsec:design phi}, we design the template with four components. Several specific examples are demonstrated in Fig.~\ref{fig:aot_example}.

\noindent\textbf{Structure.} The structure is the backbone of an ontology, referring to the essential commonality among different entities belonging to the same ontology. It is a deterministic description organized by a series of mathematical expressions. We exclusively employ differentiable operations for all the mathematical expressions involved, enabling AOTs to participate in training and optimization. This component can be seen as the constructor of a program template.

\noindent\textbf{Parameter.} 
The parameters refer to values of variables in the structure description, causing the variances of different instances belonging to the same ontology. With specific parameters, a certain instance of this template can be created. In turn, a specific target can be resolved into parameters of an ontology template.

\noindent\textbf{Affordance.} The affordance indicates the generalized knowledge of this ontology regarding where and how to interact with it. Affordances of geometric ontologies may include grasp poses, contact points for pushing, \textit{etc.} For kinematic ontologies, it usually includes the directions of force that cause movement. It is also described by mathematical expressions with certain parameters. These knowledge can provide valuable guidance to facilitate interaction.

\noindent\textbf{Renderer.} The renderer draws instances of this template in 3D space with certain data formats, like point clouds or meshes. These data can be used in many practical ways such as network training, inference and visualization.

\subsection{Discussion}
\label{subsec:aot discussion}

\noindent\textbf{Transformation to World Coordinates.} In Fig.~\ref{fig:aot_example}, we omit the common parameters of 6-dof pose (including translation/position $\boldsymbol{t}\in\mathbb{R}^3$ and rotation $\boldsymbol{r}\in\mathbf{SO}(3)$) in the world coordinate system for simplicity. Generally, it is simple to represent the pose of AOT instances in world coordinates. By representing the 6-dof pose with an affine transformation matrix $\mathbf{M}\in\mathbb{R}^{4\times4}$, each formula $F$ within the AOT structure and affordances can be reformulated from $F((x, y, z, 1)) \leq 0$ to $F(\mathbf{M}^{-1}(x, y, z, 1)) \leq 0$, thereby transforming an AOT instance to the world coordinate system from its own coordinate system. For rotationally symmetric geometries, we impose symmetry constraints on $\boldsymbol{r}$ to ensure uniqueness in the representation.

\noindent\textbf{Efforts on Defining AOT.} One of the benefits of our concept-level approach driven by AOT is that it only needs to create AOTs rather than collect real training data, and the efforts of creating AOTs are considerably low. In Tab.~\ref{tab:effort}, we compare with building CAD model~\cite{xiang2020sapien} and real-world scan~\cite{liu2022akb} datasets on both time and money costs, and our approach shows advantages in both aspects in total. This can be attributed to i) AOTs are capable of adapting to different object categories since they are developed to describe generalized concepts, and therefore a small amount of AOTs can cover the conceptual knowledge on many everyday objects, ii) new AOTs can be easily expanded in an inheritance fashion by composing existing ones and iii) a single AOT can be used to effortlessly create an infinite number of diverse synthetic data by rendering instances with different parameters. 

\begin{table}
\begin{center}
 \begin{tabular}{c||c|c||c}
 \hline
    & CAD & Scan & AOT \\
  \hline
   $T$ (min/unit) & $>120$ & 20 & $\sim 60$ \\ 
   $C_L$ (\$/unit) & $>100$ & 3 & $\sim 10$ \\
   $C_D$ (\$) & - & $\sim 1000$ & - \\
   $C_O$ (\$) & - & \checkmark & - \\
   \hline
   N (unit/category) & $\sim 30$ & $\sim 30$ & $\sim 4$ \\ 
  \hline
 \end{tabular}
  \caption{Budget comparison between building a CAD model dataset, a real-world scan dataset and analytic ontology templates. $T, C_L, C_D, C_O$ refer to time, labor cost, device cost and object cost respectively. Besides the labor costs, building a real-world scan dataset incurs extra costs of purchasing a scanner and real world objects for scanning. N refers to the number of units on average to cover an object category in terms of articulated concept discovery.}
\label{tab:effort}
\end{center}
\end{table}

\section{AOTNet}
\label{sec:AOTNet}

Taking into account the architecture and the features of AOT, we design AOTNet as a baseline to equip intelligent agents with these generalized concepts and empower them for articulated concept discovery. We first give an overview in Sec.~\ref{subsec:overview}. In Sec.~\ref{subsec:net}, we demonstrate the detailed architecture of AOTNet. Finally, the implementation details are presented in Sec.~\ref{subsec:net details}.

\subsection{Overview}
\label{subsec:overview}
\noindent\textbf{Task setting.} We study articulated concept discovery in the context of a task that given the initial and final status of a novel articulated object, an agent is asked to shift it from the initial state to the final state through interaction. This requires an intelligent system to possess the capability to comprehend the geometric and kinematic structure of an articulated part and also proper ways to interact with it, which are important topics in both the robotics and the computer vision communities. We develop AOTNet in a challenging setting that the status of the object only includes raw point clouds without any other information.

\noindent\textbf{Pipeline.} Our design aims to leverage the features of AOT and demonstrate the benefits of the AOT-driven approach. The workflow is twofold, including ontology discovery and subsequent affordance grounding (see Fig.~\ref{fig:workflow}).

As the task mainly takes into account the articulated part that undergoes state changes, we begin the workflow by discovering the geometric and kinematic ontologies on this part. The discovery process finds qualified AOT instances to capture the conceptual essence of the raw point clouds, involving two steps: ontology identification and parameter estimation. Leveraging the renderer, neural networks in these steps can be trained with synthetic data generated by rendering instances with various parameters. 

After ontology discovery, the conceptual essence of the target is captured by AOT instances with analytic structures and parameter information. According to the analytic results, we can further accurately ground the affordance knowledge on the object through mathematical calculations. In this manner, an intelligent agent is able to perform appropriate interaction guided by the knowledge.

\noindent\textbf{Discussion.} Our AOT-driven approach has strong benefits in understanding and interacting with articulated objects from three perspectives. First, AOTNet works and generalizes well across novel categories without relying on real training data, benefiting from the generalization capability of AOTs at the conceptual level. In comparison, it is necessary for a conventional object-level approach to first prepare abundant high quality articulated data for training, while acquiring, annotating, and comprehensively covering the various categories of articulated objects poses significant challenges. Second, our approach can provide analytic articulation information via AOT instances with mathematical structure and specific parameters. This also improves the interpretability of our approach. Third, affordance knowledge can be accurately grounded which brings valuable guidance for accurate and controllable interaction.

\subsection{Architecture}
\label{subsec:net}

We first introduce ontology identification and parameter estimation in the ontology discovery part, and then discuss about how grounded affordances facilitate the interaction. As the main purpose of AOTNet is to provide a baseline to demonstrate the feasibility and effectiveness of the AOT-driven paradigm, we do not delve into complex network designs. Experiments will show that promising performance can still be achieved even with a simple design.

\noindent\textbf{Ontology Identification.} Ontology identification is first introduced to identify which ontology template the input raw point clouds belong to with a classifier. The architecture of the classifier is a point cloud encoder followed by an MLP classifier. The input for geometric ontology is a single point cloud while that for kinematic ontology is a pair of point clouds at both initial and final states since at least two frames are needed to determine a dynamic process. Therefore, for the kinematic ontology, we first encode each point cloud into features separately, and then concatenate the features together to feed them into the MLP for classification.

\noindent\textbf{Parameter Estimation.} This step estimates the parameter values of an AOT to which the input belongs. 

Geometric AOTs encompass basic ones and composite ones, where the latter are formed by combining the basic ones. Since the number of parameters of composite ones increases linearly with the number of basic ones composing them, it is difficult to directly regress all the parameters of a complex composite AOT. To this end, we perform parameter estimation of geometric AOTs at the basic ontology level. Particularly, an estimator for a geometric AOT consists of four components: i) a point cloud encoder $E$ to encode the input point cloud $\mathcal{P}$ into deep features, ii) a latent code $\mathcal{R}_w$ for the whole AOT represented by learnable parameters, iii) a set of latent codes $\{\mathcal{R}_i|i=1,2,...\}$ for each basic ontology composing the AOT represented by learnable parameters, and iv) a set of MLPs $\{\operatorname{MLP}_i|i=1,2,...\}$ to regress the parameters $\{P_i|i=1,2,...\}$ of each basic ontology by $P_i=\operatorname{MLP}_i([E(\mathcal{P}),\mathcal{R}_w,\mathcal{R}_i])$, where $[\cdot]$ denotes concatenation. The set of parameters $\{P_i|i=1,2,...\}$ is merged together to produce the final estimation. Note that $i=1$ if the whole AOT is a basic one. Some parameters of basic ontology are discarded to meet the constraints in the whole AOT. To train the estimator, apart from the MSE loss on each parameter, we also introduce a Point2Mesh loss~\cite{p2mf_loss} between the mesh of the AOT instance and the input point cloud. The utilization of the Point2Mesh loss greatly benefits from the differentiable feature of AOT, meaning that the process of rendering the mesh of an AOT instance with certain parameters is differentiable. As a result, this loss can be backpropagated to the estimated parameters, enabling the optimization of the entire estimator. 

To estimate the parameters of kinematic AOTs, we use the same network architecture as the classifier and modify its output to the kinematic parameters. Particularly, the output for prismatic is its direction, and that for revolute is the axis direction and the position of the pivot point. We use cosine distance as the axis alignment loss and L2 distance to the ground truth axis as the pivot loss.

Optionally, when full-surrounding point clouds are available through multi-view observations or point cloud completion, we can optimize the estimated parameters of AOT with the aforementioned Point2Mesh loss for better geometry ontology discovery. Furthermore, the mesh rendered by AOT can be deformed to better fit the object point cloud and obtain more refined geometric details according to algorithms like \cite{hanocka2020point2mesh,pytorch3d_deformation}. Nevertheless, our experiments have shown that AOTNet can achieve good results without performing such operations.

\noindent\textbf{Affordance Grounding and Interaction.}
As mentioned above, we have discovered the descriptions of the conceptual essence of the input data, including both geometric and kinematic concepts as well as the parameters. At this point, corresponding affordances in the AOTs can be grounded mathematically. According to this information, interactions can be conducted on the object to effectively change the state of the articulation as required. In our implementation, we just use a simple interaction strategy that first holds the articulated part according to the geometric affordances and then applies force along the direction according to the kinematic affordances.

\subsection{Implementation Details}
\label{subsec:net details}

\noindent\textbf{Training Data Preparation.} All the neural networks involved in AOTNet are trained with synthetic point clouds of AOT instances with different parameters, which are generated by the renderer. Particularly, the renderer samples points in 3D space according to the mathematical expressions of the structure to acquire the point cloud of a geometric ontology. For the kinematic ontology, the renderer produces a pair of point clouds, which represent the states of a geometric ontology before and after the movement. Some additional noise and corruptions are introduced to the point clouds for data augmentation. 

\noindent\textbf{Networks.} The point cloud encoders used in both ontology identification and parameter estimation are transformer encoders~\cite{yu2021pointbert}, which extract 128 groups of points with size 32 from the input with 2048 points, and send them into a standard transformer encoder with 12 6-headed attention layers. All MLPs used in our experiments are triple-layered with ReLU activation.

\begin{figure*}[tb!]
\centering
\includegraphics[width=1.0\textwidth]{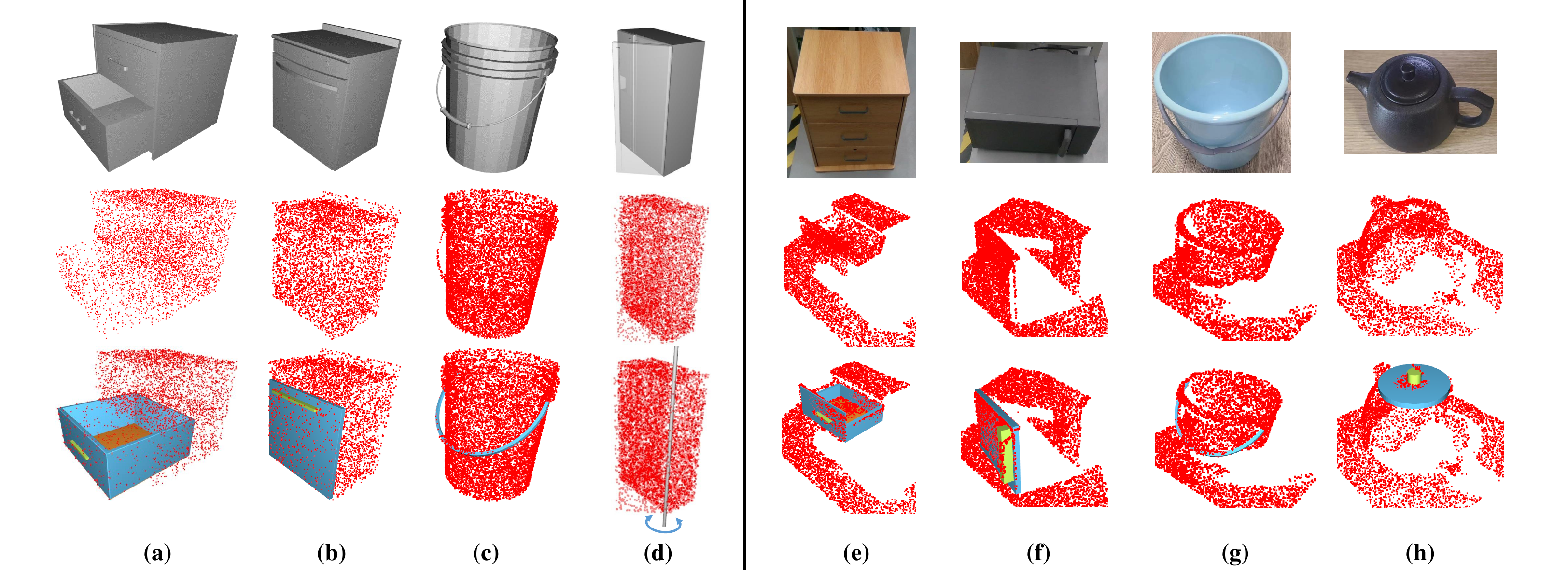}
\caption{Visualization of ontology discovery results for real-world objects in a simulation environment (a-d) and the physical world (e-h). The first row shows the target objects. The second row gives the input point clouds of the objects. The third row shows the discovery results of the actionable part with AOTNet.}
\label{fig:recon_example}
\end{figure*}

\section{Experiments}
\label{sec:exp}

We evaluate the full AOTNet with a closed-loop validation through the success rate of interaction on the proposed task, \textit{i.e.} changing an articulated object from its initial state to a target final state. The success rate can reveal the quality of articulated concept discovery, including ontology discovery and affordance grounding. More experimental details, results and analysis can be found in the supplementary materials.

\noindent\textbf{Experiment Settings.}
We conduct our experiments in SAPIEN~\cite{xiang2020sapien} environment and follow \cite{mo2021where2act} for the environment setting. We use a flying Franka Panda gripper as the agent. To generate the input partial point cloud scans, we mount an RGB-D camera with known intrinsic parameters 5-unit-length away pointing to the center of the target object.

To satisfy the compatibility with a single-gripper, we evaluate on a total of 17 specific interaction tasks involving 15 representative object categories chosen from PartNet-Mobility Dataset, including turning faucets, opening table doors, flipping bucket handles, \textit{etc.} In total, 653 objects with 1152 articulation structures are used as test samples. A full list of the specific tasks is provided in the supplementary materials. For each specific task, we set the initial state of the target joint as closed (0\% of the joint range), and the final state as fully open (100\% of the joint range). In practice, we consider a task successful if more than 80\% of the total moving range is achieved.

\noindent\textbf{Baselines.} To compare with the interaction strategy of AOTNet, we adopt two popular reinforcement learning approaches, Soft Actor-Critic (SAC)~\cite{haarnoja2018soft} and Truncated Quantile Critics (TQC)~\cite{kuznetsov2020controlling}, as baselines. For both RL approaches, we provide three different state representations: 
\begin{itemize}
    \item Raw: raw input point cloud of the target.
    \item Raw + JP: information in Raw, and both ground-truth joint specifications and 6-dof poses of the articulation structure.
    \item AOT: AOT instances (template type and corresponding parameters) discovered on the target by AOTNet, including both geometric and kinematic ones.
\end{itemize}
Note that the pose, velocity and gripper contact of the agent are the common states shared among the three settings. The training samples for SAC/TQC are selected from PartNet-Mobility assets except those in test samples. We also provide the \textit{Human} evaluation result as a reference, in which we first invite ten college students to discover the AOTs on the test cases according to their cognition and then use the interaction strategy in AOTNet with these manually discovered AOTs for interaction.

\begin{table*}
\begin{center}
\begin{tabular}{c||c|c||ccc|ccc}
\hline
\multirow{2}{*}{Method} & \multirow{2}{*}{Human} & \multirow{2}{*}{AOTNet} & \multicolumn{3}{c|}{SAC}                                         & \multicolumn{3}{c}{TQC}                                         \\ \cline{4-9} 
                        &                        &                         & \multicolumn{1}{c|}{Raw}  & \multicolumn{1}{c|}{Raw + JP} & AOT  & \multicolumn{1}{c|}{RAW}  & \multicolumn{1}{c|}{Raw + JP} & AOT  \\ \hline
Avg. Succ. Rate         & 97.5                   & 63.7                    & \multicolumn{1}{c|}{19.6} & \multicolumn{1}{c|}{41.5}     & 34.3 & \multicolumn{1}{c|}{22.6} & \multicolumn{1}{c|}{46.4}     & 39.0 \\ \hline
\end{tabular}
\caption{Success rate of articulated object interaction.}
\label{tab:manipulation}
\end{center}
\end{table*}

\noindent\textbf{Main Results.} Tab.~\ref{tab:manipulation} gives the average success rate of our approach and other baseline methods. The success rate of human evaluation is very close to 100\%, demonstrating that AOTs can adequately reflect the conceptual knowledge of structure and affordance on articulated objects when they are discovered based on human cognition. For AOTNet, it achieves a remarkable result of 63.7\% success rate which greatly outperforms RL-based methods, indicating that the ontology discovery and affordance grounding in AOTNet enjoy good accuracy. Given that AOTNet does not involve real training data and all these test articulated objects are novel for AOTNet, this also demonstrates the superiority of learning at the conceptual level. Fig.~\ref{fig:recon_example} shows visualizations of ontology discovery, and demonstration videos are provided in the supplementary material for further qualitative analysis, including real-world experiments.

\noindent\textbf{Contribution of Affordance Information.} The full AOTNet performs interaction according to grounded affordances, while the RL-based approaches explore the interaction strategies according to the input state representations from actual interactions. According to Tab.~\ref{tab:manipulation}, AOTNet performs better than all the RL settings, especially Raw+JP which introduces joint and pose ground truths. This ablation experiment demonstrates that the affordance knowledge is well grounded, and provides valuable guidance for accurate interaction. In comparison, it is relatively more difficult to explore proper affordances and corresponding interaction strategies with RL from the input representations.

\noindent\textbf{Contribution of Analytic Articulation Information.} The RL results in Tab.~\ref{tab:manipulation} demonstrate that the performance of the three settings shows a similar trend for both SAC and TQC. The Raw setting does not perform well compared with the others considering there is no joint and 6-dof pose information. With ground truths as additional state representations, Raw+JP achieves the best performance among the three. The result of AOT setting is much higher than Raw, indicating one of the benefits of an AOT-driven approach that an analytic description can be given for the raw input, which enables RL agents to better leverage the state representations to learn interaction strategies.

\section{Related Work}
\label{sec:rwfw}
\noindent\textbf{Conceptual Object Understanding in Human Cognition.} Researchers in cognitive science have been studying visual object understanding for several decades~\cite{humphreys1999objects,ullman2000high,palmeri2004visual,biederman1987recognition,hummel1992dynamic}. Their investigations have aimed to unravel the intricate processes involved in perceiving, recognizing, and comprehending objects within the human mind, wherein they have discovered the significant role of conceptual knowledge~\cite{biederman1987recognition,habel2006abstract,palmeri2004visual,rosch1975cognitive}. For example, Biederman~\cite{biederman1987recognition} has found that the perceptual recognition of objects can be conceptualized to be a process in which an object is segmented into an arrangement of simple geometric components, such as blocks, cylinders, wedges, and cones. Compelling evidence provided in \cite{palmeri2004visual,dixon2002role,goldstone1998reuniting} also indicates a strong relation between perception and conceptual knowledge. Studies on infants~\cite{carey2001infants,scholl1999explaining} further reveals the pivotal significance of conceptual knowledge in object understanding, since infants are much less susceptible to empirical factors. These findings establish a possible way for machine intelligence to understand objects at the conceptual level, but little previous work in the computer vision community has paid attention to this area.

\noindent\textbf{Articulated Object Understanding.} Articulation structure is an important topic in both vision and robotics community and has been investigated by many researchers. Many articulated object assets~\cite{xiang2020sapien,liu2022akb,martin2019rbo,calli2015ycb} have been proposed in recent years. Thanks to previous work ShapeNet~\cite{chang2015shapenet} that collects a large amount of CAD object models and PartNet~\cite{mo2019partnet} that gives hierarchical part semantic segmentation annotations on a subset of ShapeNet, PartNet-Mobility~\cite{xiang2020sapien} and Shape2Motion~\cite{wang2019shape2motion} further label the joint information of CAD models in PartNet for articulation research. On the other hand, some researches~\cite{liu2022akb} aim to collect assets via real-world scanning to build articulated object datasets. 

Based on these datasets, articulated objects have been studied from many aspects. From the perception aspect, current works concentrate on multiple areas such as recognition~\cite{zeng2021visual,jain2021screwnet}, part segmentation~\cite{yi2018deep}, pose estimation~\cite{li2020category,liu2022toward} and tracking~\cite{heppert2022category, weng2021captra}. For example, Yi et al.~\cite{yi2018deep} develop a network to co-segment the input objects into their articulated parts according to a 3D CAD model similar to the input. Li et al.~\cite{li2020category} propose a normalized coordinate space to estimate 6D pose and joint state for articulated objects. There are also many researches on discovering how to interact with articulated objects~\cite{katz2008manipulating, xiang2020sapien, mo2021where2act, liu2022akb, wang2022adaafford, geng2023gapartnet, ling2024articulated}. For example, Mo et al.~\cite{mo2021where2act,wang2022adaafford} extract highly localized actionable information for articulated objects with movable parts. Still, most of current approaches work on object or part level and heavily rely on high-quality 3D data for training. In comparison, our AOT-driven paradigm demonstrates promising performance at conceptual level and needs no real training data for its generalization capability.

\section{Conclusion}
\label{sec:conclusion}

In this paper, we introduce Analytic Ontology Templates as the description for generalized concepts, in order to enable machine intelligence to appropriately perceive, comprehend and interact with articulated objects at the conceptual level, especially for those in novel categories. Our main contributions are as follows. First, we propose Analytic Ontology Template as a parameterized and differentiable template description of a generalized conceptual ontology, and affordances can be numerically described in the template for accurate and controllable interactions. Second, an AOT-driven pipeline called AOTNet is designed accordingly to equip intelligent agents with these concepts, and then empower the agents to effectively discover the conceptual knowledge of structure and affordance on objects. Third, we comprehensively evaluate the effectiveness of AOT and AOTNet on hundreds of articulated objects across a wide range of categories in terms of articulated concept discovery. The experiments suggest the benefits of the AOT-driven paradigm and the possibility for machine intelligence to mimic the human capability of learning at the conceptual level.

\bibliography{aaai25}

\begin{thebibliography}{38}
\providecommand{\natexlab}[1]{#1}

\bibitem[{Biederman(1987)}]{biederman1987recognition}
Biederman, I. 1987.
\newblock Recognition-by-components: a theory of human image understanding.
\newblock \emph{Psychological review}, 94(2): 115.

\bibitem[{Calli et~al.(2015)Calli, Singh, Walsman, Srinivasa, Abbeel, and Dollar}]{calli2015ycb}
Calli, B.; Singh, A.; Walsman, A.; Srinivasa, S.; Abbeel, P.; and Dollar, A.~M. 2015.
\newblock The ycb object and model set: Towards common benchmarks for manipulation research.
\newblock In \emph{2015 international conference on advanced robotics (ICAR)}, 510--517. IEEE.

\bibitem[{Carey and Xu(2001)}]{carey2001infants}
Carey, S.; and Xu, F. 2001.
\newblock Infants' knowledge of objects: Beyond object files and object tracking.
\newblock \emph{Cognition}, 80(1-2): 179--213.

\bibitem[{Chang et~al.(2015)Chang, Funkhouser, Guibas, Hanrahan, Huang, Li, Savarese, Savva, Song, Su et~al.}]{chang2015shapenet}
Chang, A.~X.; Funkhouser, T.; Guibas, L.; Hanrahan, P.; Huang, Q.; Li, Z.; Savarese, S.; Savva, M.; Song, S.; Su, H.; et~al. 2015.
\newblock Shapenet: An information-rich 3d model repository.
\newblock \emph{arXiv preprint arXiv:1512.03012}.

\bibitem[{Dixon et~al.(2002)Dixon, Desmarais, Gojmerac, Schweizer, and Bub}]{dixon2002role}
Dixon, M.~J.; Desmarais, G.; Gojmerac, C.; Schweizer, T.~A.; and Bub, D.~N. 2002.
\newblock The role of premorbid expertise on object identification in a patient with category-specific visual agnosia.
\newblock \emph{Cognitive Neuropsychology}, 19(5): 401--419.

\bibitem[{Geng et~al.(2023)Geng, Xu, Zhao, Xu, Yi, Huang, and Wang}]{geng2023gapartnet}
Geng, H.; Xu, H.; Zhao, C.; Xu, C.; Yi, L.; Huang, S.; and Wang, H. 2023.
\newblock Gapartnet: Cross-category domain-generalizable object perception and manipulation via generalizable and actionable parts.
\newblock In \emph{Proceedings of the IEEE/CVF Conference on Computer Vision and Pattern Recognition}, 7081--7091.

\bibitem[{Goldstone and Barsalou(1998)}]{goldstone1998reuniting}
Goldstone, R.~L.; and Barsalou, L.~W. 1998.
\newblock Reuniting perception and conception.
\newblock \emph{Cognition}, 65(2-3): 231--262.

\bibitem[{Haarnoja et~al.(2018)Haarnoja, Zhou, Abbeel, and Levine}]{haarnoja2018soft}
Haarnoja, T.; Zhou, A.; Abbeel, P.; and Levine, S. 2018.
\newblock Soft actor-critic: Off-policy maximum entropy deep reinforcement learning with a stochastic actor.
\newblock In \emph{International conference on machine learning}, 1861--1870. PMLR.

\bibitem[{Habel and Eschenbach(2006)}]{habel2006abstract}
Habel, C.; and Eschenbach, C. 2006.
\newblock Abstract structures in spatial cognition.
\newblock \emph{Foundations of Computer Science: Potential—Theory—Cognition}, 369--378.

\bibitem[{Hanocka et~al.(2020)Hanocka, Metzer, Giryes, and Cohen-Or}]{hanocka2020point2mesh}
Hanocka, R.; Metzer, G.; Giryes, R.; and Cohen-Or, D. 2020.
\newblock Point2mesh: A self-prior for deformable meshes.
\newblock \emph{arXiv preprint arXiv:2005.11084}.

\bibitem[{Heppert et~al.(2022)Heppert, Migimatsu, Yi, Chen, and Bohg}]{heppert2022category}
Heppert, N.; Migimatsu, T.; Yi, B.; Chen, C.; and Bohg, J. 2022.
\newblock Category-Independent Articulated Object Tracking with Factor Graphs.
\newblock In \emph{2022 IEEE/RSJ International Conference on Intelligent Robots and Systems (IROS)}, 3800--3807. IEEE.

\bibitem[{Hummel and Biederman(1992)}]{hummel1992dynamic}
Hummel, J.~E.; and Biederman, I. 1992.
\newblock Dynamic binding in a neural network for shape recognition.
\newblock \emph{Psychological review}, 99(3): 480.

\bibitem[{Humphreys, Price, and Riddoch(1999)}]{humphreys1999objects}
Humphreys, G.~W.; Price, C.~J.; and Riddoch, M.~J. 1999.
\newblock From objects to names: A cognitive neuroscience approach.
\newblock \emph{Psychological research}, 62: 118--130.

\bibitem[{Jain et~al.(2021)Jain, Lioutikov, Chuck, and Niekum}]{jain2021screwnet}
Jain, A.; Lioutikov, R.; Chuck, C.; and Niekum, S. 2021.
\newblock Screwnet: Category-independent articulation model estimation from depth images using screw theory.
\newblock In \emph{2021 IEEE International Conference on Robotics and Automation (ICRA)}, 13670--13677. IEEE.

\bibitem[{Katz and Brock(2008)}]{katz2008manipulating}
Katz, D.; and Brock, O. 2008.
\newblock Manipulating articulated objects with interactive perception.
\newblock In \emph{2008 IEEE International Conference on Robotics and Automation}, 272--277. IEEE.

\bibitem[{Kuznetsov et~al.(2020)Kuznetsov, Shvechikov, Grishin, and Vetrov}]{kuznetsov2020controlling}
Kuznetsov, A.; Shvechikov, P.; Grishin, A.; and Vetrov, D. 2020.
\newblock Controlling overestimation bias with truncated mixture of continuous distributional quantile critics.
\newblock In \emph{International Conference on Machine Learning}, 5556--5566. PMLR.

\bibitem[{Leslie et~al.(1998)Leslie, Xu, Tremoulet, and Scholl}]{leslie1998indexing}
Leslie, A.~M.; Xu, F.; Tremoulet, P.~D.; and Scholl, B.~J. 1998.
\newblock Indexing and the object concept: developingwhat'andwhere'systems.
\newblock \emph{Trends in cognitive sciences}, 2(1): 10--18.

\bibitem[{Li et~al.(2020)Li, Wang, Yi, Guibas, Abbott, and Song}]{li2020category}
Li, X.; Wang, H.; Yi, L.; Guibas, L.~J.; Abbott, A.~L.; and Song, S. 2020.
\newblock Category-level articulated object pose estimation.
\newblock In \emph{Proceedings of the IEEE/CVF Conference on Computer Vision and Pattern Recognition}, 3706--3715.

\bibitem[{Ling et~al.(2024)Ling, Wang, Wu, Wu, Zhuang, Xu, Li, Liu, and Dong}]{ling2024articulated}
Ling, S.; Wang, Y.; Wu, R.; Wu, S.; Zhuang, Y.; Xu, T.; Li, Y.; Liu, C.; and Dong, H. 2024.
\newblock Articulated object manipulation with coarse-to-fine affordance for mitigating the effect of point cloud noise.
\newblock In \emph{2024 IEEE International Conference on Robotics and Automation (ICRA)}, 10895--10901. IEEE.

\bibitem[{Liu et~al.(2022{\natexlab{a}})Liu, Xu, Fu, Qian, Yu, Han, and Lu}]{liu2022akb}
Liu, L.; Xu, W.; Fu, H.; Qian, S.; Yu, Q.; Han, Y.; and Lu, C. 2022{\natexlab{a}}.
\newblock AKB-48: a real-world articulated object knowledge base.
\newblock In \emph{Proceedings of the IEEE/CVF Conference on Computer Vision and Pattern Recognition}, 14809--14818.

\bibitem[{Liu et~al.(2022{\natexlab{b}})Liu, Xue, Xu, Fu, and Lu}]{liu2022toward}
Liu, L.; Xue, H.; Xu, W.; Fu, H.; and Lu, C. 2022{\natexlab{b}}.
\newblock Toward real-world category-level articulation pose estimation.
\newblock \emph{IEEE Transactions on Image Processing}, 31: 1072--1083.

\bibitem[{Mart{\'\i}n-Mart{\'\i}n, Eppner, and Brock(2019)}]{martin2019rbo}
Mart{\'\i}n-Mart{\'\i}n, R.; Eppner, C.; and Brock, O. 2019.
\newblock The RBO dataset of articulated objects and interactions.
\newblock \emph{The International Journal of Robotics Research}, 38(9): 1013--1019.

\bibitem[{Mo et~al.(2021)Mo, Guibas, Mukadam, Gupta, and Tulsiani}]{mo2021where2act}
Mo, K.; Guibas, L.~J.; Mukadam, M.; Gupta, A.; and Tulsiani, S. 2021.
\newblock Where2act: From pixels to actions for articulated 3d objects.
\newblock In \emph{Proceedings of the IEEE/CVF International Conference on Computer Vision}, 6813--6823.

\bibitem[{Mo et~al.(2019)Mo, Zhu, Chang, Yi, Tripathi, Guibas, and Su}]{mo2019partnet}
Mo, K.; Zhu, S.; Chang, A.~X.; Yi, L.; Tripathi, S.; Guibas, L.~J.; and Su, H. 2019.
\newblock Partnet: A large-scale benchmark for fine-grained and hierarchical part-level 3d object understanding.
\newblock In \emph{Proceedings of the IEEE/CVF conference on computer vision and pattern recognition}, 909--918.

\bibitem[{Palmeri and Gauthier(2004)}]{palmeri2004visual}
Palmeri, T.~J.; and Gauthier, I. 2004.
\newblock Visual object understanding.
\newblock \emph{Nature Reviews Neuroscience}, 5(4): 291--303.

\bibitem[{Piaget(1955)}]{piaget1955child}
Piaget, J. 1955.
\newblock \emph{The child's construction of reality}.
\newblock London.

\bibitem[{PyTorch3D(2023{\natexlab{a}})}]{pytorch3d_deformation}
PyTorch3D. 2023{\natexlab{a}}.
\newblock pytorch3d\_deformation.
\newblock \url{https://github.com/facebookresearch/pytorch3d/blob/main/docs/tutorials/deform_source_mesh_to_target_mesh.ipynb}.

\bibitem[{PyTorch3D(2023{\natexlab{b}})}]{p2mf_loss}
PyTorch3D. 2023{\natexlab{b}}.
\newblock pytorch3d.loss.point\_mesh\_face\_distance.
\newblock \url{https://pytorch3d.readthedocs.io/en/latest/modules/loss.html##pytorch3d.loss.point_mesh_face_distance}.

\bibitem[{Rosch(1975)}]{rosch1975cognitive}
Rosch, E. 1975.
\newblock Cognitive representations of semantic categories.
\newblock \emph{Journal of experimental psychology: General}, 104(3): 192.

\bibitem[{Scholl and Leslie(1999)}]{scholl1999explaining}
Scholl, B.; and Leslie, A. 1999.
\newblock Explaining the infant’s object concept.
\newblock \emph{What is cognitive science}, 26--73.

\bibitem[{Ullman(2000)}]{ullman2000high}
Ullman, S. 2000.
\newblock \emph{High-level vision: Object recognition and visual cognition}.
\newblock MIT press.

\bibitem[{Wang et~al.(2019)Wang, Zhou, Shi, Chen, Zhao, and Xu}]{wang2019shape2motion}
Wang, X.; Zhou, B.; Shi, Y.; Chen, X.; Zhao, Q.; and Xu, K. 2019.
\newblock Shape2motion: Joint analysis of motion parts and attributes from 3d shapes.
\newblock In \emph{Proceedings of the IEEE/CVF Conference on Computer Vision and Pattern Recognition}, 8876--8884.

\bibitem[{Wang et~al.(2022)Wang, Wu, Mo, Ke, Fan, Guibas, and Dong}]{wang2022adaafford}
Wang, Y.; Wu, R.; Mo, K.; Ke, J.; Fan, Q.; Guibas, L.~J.; and Dong, H. 2022.
\newblock Adaafford: Learning to adapt manipulation affordance for 3d articulated objects via few-shot interactions.
\newblock In \emph{Computer Vision--ECCV 2022: 17th European Conference, Tel Aviv, Israel, October 23--27, 2022, Proceedings, Part XXIX}, 90--107. Springer.

\bibitem[{Weng et~al.(2021)Weng, Wang, Zhou, Qin, Duan, Fan, Chen, Su, and Guibas}]{weng2021captra}
Weng, Y.; Wang, H.; Zhou, Q.; Qin, Y.; Duan, Y.; Fan, Q.; Chen, B.; Su, H.; and Guibas, L.~J. 2021.
\newblock Captra: Category-level pose tracking for rigid and articulated objects from point clouds.
\newblock In \emph{Proceedings of the IEEE/CVF International Conference on Computer Vision}, 13209--13218.

\bibitem[{Xiang et~al.(2020)Xiang, Qin, Mo, Xia, Zhu, Liu, Liu, Jiang, Yuan, Wang et~al.}]{xiang2020sapien}
Xiang, F.; Qin, Y.; Mo, K.; Xia, Y.; Zhu, H.; Liu, F.; Liu, M.; Jiang, H.; Yuan, Y.; Wang, H.; et~al. 2020.
\newblock Sapien: A simulated part-based interactive environment.
\newblock In \emph{Proceedings of the IEEE/CVF Conference on Computer Vision and Pattern Recognition}, 11097--11107.

\bibitem[{Yi et~al.(2018)Yi, Huang, Liu, Kalogerakis, Su, and Guibas}]{yi2018deep}
Yi, L.; Huang, H.; Liu, D.; Kalogerakis, E.; Su, H.; and Guibas, L. 2018.
\newblock Deep part induction from articulated object pairs.
\newblock \emph{arXiv preprint arXiv:1809.07417}.

\bibitem[{Yu et~al.(2022)Yu, Tang, Rao, Huang, Zhou, and Lu}]{yu2021pointbert}
Yu, X.; Tang, L.; Rao, Y.; Huang, T.; Zhou, J.; and Lu, J. 2022.
\newblock Point-BERT: Pre-Training 3D Point Cloud Transformers with Masked Point Modeling.
\newblock In \emph{Proceedings of the IEEE Conference on Computer Vision and Pattern Recognition (CVPR)}.

\bibitem[{Zeng et~al.(2021)Zeng, Lee, Liang, and Kroemer}]{zeng2021visual}
Zeng, V.; Lee, T.~E.; Liang, J.; and Kroemer, O. 2021.
\newblock Visual identification of articulated object parts.
\newblock In \emph{2021 IEEE/RSJ International Conference on Intelligent Robots and Systems (IROS)}, 2443--2450. IEEE.

\end{thebibliography}

\end{document}